\def\year{2020}\relax
\documentclass[letterpaper]{article} % DO NOT CHANGE THIS
\usepackage{aaai20}  % DO NOT CHANGE THIS
\usepackage{times}  % DO NOT CHANGE THIS
\usepackage{helvet} % DO NOT CHANGE THIS
\usepackage{courier}  % DO NOT CHANGE THIS
\usepackage[hyphens]{url}  % DO NOT CHANGE THIS
\usepackage{graphicx} % DO NOT CHANGE THIS
\usepackage{comment}
\usepackage{amsmath}
\usepackage{graphicx}  % DO NOT CHANGE THIS
\title{A Preliminary Approach for Learning Relational Policies for the Management of Critically Ill Children}
\author{Michael A. Skinner,\textsuperscript{\rm 1,2} Lakshmi Raman, \textsuperscript{\rm 2} Neel Shah, \textsuperscript{\rm 2} Abdelaziz Farhat, \textsuperscript{\rm 2}Sriraam Natarajan, \textsuperscript{\rm 1}\\ % All authors must be in the same font size and format. Use \Large and \textbf to achieve this result when breaking a line
\textsuperscript{\rm 1}University of Texas at Dallas, Dallas TX\\ 
\textsuperscript{\rm 2}University of Texas Southwestern Medical Center, Dallas TX%If you have multiple authors and multiple affiliations
}
\begin{document}

\maketitle

%\begin{abstract}
% This is samplepaper.tex, a sample chapter demonstrating the
% LLNCS macro package for Springer Computer Science proceedings;
% Version 2.20 of 2017/10/04
%
%\documentclass[runningheads]{llncs}
%\usepackage[
%backend=biber,
%style=alphabetic,
%sorting=ynt
%]{biblatex}
%\addbibresource{bibliography.bib}
%
%\usepackage{graphicx}
%\usepackage{wrapfig}
%\usepackage[dvipsnames]{xcolor}
%\usepackage{xcolor}
%\usepackage{array}
%\usepackage{verbatim}
% Used for displaying a sample figure. If possible, figure files should
% be included in EPS format.
%
% If you use the hyperref package, please uncomment the following line
% to display URLs in blue roman font according to Springer's eBook style:
% \renewcommand\UrlFont{\color{blue}\rmfamily}
\newcommand{\sn}[1]{\textcolor{Red}{[#1 \textsc{--Sriraam}]}}
\newcommand{\ms}[1]{\textcolor{Blue}{[#1 \textsc{--Mike}]}}
%\begin{document}
%
 % treated with extracorporeal membrane oxygenation}
%
%\titlerunning{Abbreviated paper title}
% If the paper title is too long for the running head, you can set
% an abbreviated paper title here
%
%\author{Michael A. Skinner\inst{1,2} \and
%Lakshmi Raman\inst{2} \and
%Neel Shah\inst{2} \and
%Abdelaziz Farhat\inst{2} \and
%Sriraam Natarajan\inst{1}}
%
%\authorrunning{M. Skinner et al.}
% First names are abbreviated in the running head.
% If there are more than two authors, 'et al.' is used.
%
%\institute{University of Texas at Dallas, Dallas TX \and University of Texas Southwestern Medical Center, Dallas TX}
%
%\maketitle              % typeset the header of the contribution
%
%\email{\{michael.skinner, sriraam.natarajan\}@utdallas.edu,\\ \{lakshmi.raman, neel.shah, abdelaziz.farhat\}@utsouthwestern.edu}
\begin{abstract}
The increased use of electronic health records has made possible the automated extraction of medical policies from patient records to aid in the development of clinical decision support systems. We adapted a boosted Statistical Relational Learning (SRL) framework to learn probabilistic rules from clinical hospital records for the management of physiologic parameters of children with severe cardiac or respiratory failure who were managed with extracorporeal membrane oxygenation. In this preliminary study, the results were promising. In particular, the algorithm returned logic rules for medical actions that are consistent with medical reasoning.

%\keywords{Probabilistic logic learning  \and medical policy.}
\end{abstract}
\section{Introduction}
The ability to automatically learn physician actions from electronic health records (EHR) could contribute to improved health care in a number of ways. For example, we could automatically discover optimal policies~\footnote{We follow the conventional reinforcement learning definition of a policy as a mapping from states to actions} for managing particular diseases. Moreover, an optimal policy, once discovered, could be compared to a patient's actual clinical course; if there is a deviation, physicians could be provided with suggestions for care. Finally, the ability to extract medical polices from EHRs would enable predictions of patient prognosis and outcomes.

In this work-in-progress, we investigate use of a boosted SRL framework to elicit weighted first-order logic clauses mapping the values of a set of physiologic parameters to physician actions in critically ill patients with respiratory or cardiac failure. We extracted the information from clinical trajectories documented in the EHR. The goal of this work is to explore the use of such frameworks in this challenging medical task.~\footnote{This work was presented in part at the 2019 Probabilistic Logic Programming Workshop, Las Cruces NM.}

%The paper is organized as follows. The motivation for learning the clinical management policy is discussed in Section 2. In Section 3, we describe the patients and the data, and describe our methods. Section 4 presents our results. We discuss our findings, shortcomings, and possible future work in Section 5.

\textbf{The clinical challenge -- discovering a medical policy} Unwanted variation in medical care, recognized for over forty years, remains a challenge to health care providers in nearly every specialty~\cite{westert2018medical,karimi2017national,lilot2015variability}. Differences in care are typically observed between geographic regions, and the particular practice in an area often correlates with available resources. For example in one study, investigators found a high correlation between the availability of cardiac catheterization within a locality and use of angioplasty for managing cardiac disease~\cite{brownlee2010overtreated}. It is surprising that these challenges persist, even as there has been a multiplication of published expert guideline documents for many medical conditions whose recommendations are based on well-performed prospective clinical trials~\cite{weisz2007emergence}.

To decrease variability of care, and to converge medical management around policies conforming to expert guidelines, clinical decision support systems (CDS) have been devised to render advice to clinicians as they care for patients~\cite{osheroff2007roadmap}. Such systems were initially very limited, highly dependent on manual curation, and their scope was limited to a very few medical conditions~\cite{middleton2016clinical}. However, the increased use of EHRs has stimulated the development of automated CDS systems holding promise for providing advice to health care providers in real time.

Required in an automated CDS system is the ability to monitor some aspects of the patient's clinical state, as well as the physician actions~\cite{middleton2016clinical}. Moreover, the system must possess some notion of optimal care; when clinicians deviate from the preferred management, or if unexpected events occur warranting a change in care, alerts or reminders are provided from the system. Whereas early systems used hard-coded rules to encode clinicians' knowledge about the optimal policy, there is growing interest in automatically extracting optimal care patterns by mining the EHR~\cite{ohno2016using}.

Reinforcement learning (RL) is the most commonly reported technique to extract clinical policies from medical records~\cite{komorowski2018artificial,raghu2017continuous}. Somewhat surprisingly, to our knowledge, other policy learning schemes such as imitation learning have not been reported in the medical realm. Most of the reported RL models use deep neural networks, requiring many patient records for training and a propositionalization/embedding technique that could lead to loss of information.  Moreover, these models may be difficult to interpret, complicating the identification of best medical actions for a given patient state. %\sn{Why PLM?}

A key issue when learning medical policies is that it is non-trivial to construct a vector-based representation for EHR information. There can be multiple measurements performed over varying time-scales, multiple treatments of different conditions at the same time-step and differing numbers of observations per subject. Thus, if standard machine learning methods are used to represent these complicated data, critical information may be lost.

A more natural representation that allows for modeling relational medical data is to use first-order logic. Hence, motivated by the fact that physicians generally make treatment decisions based on physical, laboratory and radiologic findings in a systematic manner through a series of (often implicit) ''if-then" decisions, we investigated the usefulness of learning medical polices as sets of probabilistic clauses learned in a statistical relational learning framework~\cite{de2008probabilistic,raedt2016statistical}.%\ms{IS it relevant that EHrs are giant relational databases?}
\section{Problem Description - ECMO patients}
%\subsection{ECMO patients}
Extracorporeal membrane oxygenation (ECMO) is a method of supporting patients with severe respiratory or cardiac failure. The technique requires placement of large cannulas in the neck or in the heart, and externally circulating the patient's blood through a system that oxygenates the blood and removes carbon dioxide. Reserved for the most critically ill of patients, mortality can be very high and even among survivors there are frequent treatment complications~\cite{lin2017extracorporeal}. 

This study used de-identified medical data abstracted from EHRs for 140 children treated at the Children's Medical Center of Dallas who survived their period of ECMO. The study was performed in accordance with an exemption granted by the University of Texas Southwestern Institutional Review Board (IRB). The time on ECMO ranged from 6 to 985 hours, averaging 174 hours. For each hour of ECMO bypass, and for from 1 to 24 hours prior to cannulation (15 hours, on average), 40 physiologic and laboratory parameters were recorded. Not every parameter was measured each hour; for example, those exclusively associated with ongoing bypass (such as pump flow) were only recorded while the child was actually undergoing ECMO support.

We chose seven physiologic parameters thought to be the most useful for managing the respiratory and hemodynamic status of patients. These are tabulated, with the units of measurement, in Table~\ref{tab:paramList}. Parameter values were each discretized into five bins; the demarcations were based on meaningful physiologic categories. Thus, for example, the range of Mean arterial pressure (MAP) values was {50,60,70,80,80.1} (mm Hg.). If the $MAP \leq 50$, then the bin was labeled 50, if $MAP> 80$, then the assigned bin was 80.1, and if $50< MAP \leq 60$, the bin assignment was 60, and so forth. Of course, the bin values and/or units are different for each parameter.
%\\
\begin{table}
\centering
%\end{wraptable}{r}{6cm}
\caption{Study parameters.}
%\\
\label{tab:paramList}
\begin{tabular}{|l|l|}
\hline
Parameter &  Units\\
\hline
Mean arterial pressure & mm Hg.\\
Heart rate & beats/min\\
Respiratory rate & breaths/min\\
pH & none\\
pO2 & mm Hg.\\
Pressure volume sensor & cm H2O\\
Measured flow & ml/kg-min\\
\hline
\end{tabular}
%\caption{Study parameters.}
%\\
\end{table}

\section{Statistical relational logic models for learning medical policies in ECMO patients}

Formally, given EHR data from a set of patients and the set of actions listed in Table~\ref{tab:actionList}, we seek to learn (parameterized) policies for specifying the appropriate medical actions to alter physiologic parameters. In other words, our aim is to learn from the data when physicians should initiate therapy to alter the parameters listed in Table~\ref{tab:paramList}.
\begin{table}
\centering
\caption{Policy actions.}\label{tab:actionList}
\begin{tabular}{|l|}
\hline
Action\\
\hline
Increase mean arterial pressure\\
Increase/decrease respiratory rate\\
Decrease heart rate\\
Increase/decrease pH\\
Increase/decease pO2\\
Increase/decrease pressure volume sensor\\
Increase/decrease measured flow\\
\hline
\end{tabular}
\end{table}

We are inspired by prior work on learning policies using SRL models~\cite{natarajan2011imitation} where (parameterized weighted logical) clauses were learned from observed trajectories. Broadly known as ''imitation learning", the key idea is to learn a distribution over actions such that the policies are as close to the observed user policy as possible. This particular setting is quite useful in cases where the reward function is difficult to specify in advance. Imitation learning algorithms directly optimize the learned policy from trajectories instead of the expected cumulative discounted reward (as in reinforcement learning); in many cases this is easier, since when we have observed trajectories, we can avoid exploration. We consider learning from observations and learn a relational policy from data. 

Our SRL learning method is based on learning a set of logical regression (TILDE) trees~\cite{blockeel1998top} in a stage-wise manner. This learning method uses an underlying Inductive Logic Programming (ILP) learner~\cite{muggleton1992inductive} to induce a set of logical clauses and then fits the weights (parameters) of these clauses. We employ the machinery of gradient-boosting~\cite{natarajan2015boosted} where differences between observed and predicted probabilities are computed as gradients for the training examples and TILDE trees are learned at each step to fit these gradients. For more details, we refer to our previous work~\cite{natarajan2015boosted}.

Recall that an ILP algorithm accepts a set of facts, sets of positive and negative examples of the concepts to learn, returning logic programs defining the learned concepts. To learn the concepts listed in Table~\ref{tab:actionList}, we include as facts the values of each parameter for each subject for each hour; for instance, "map(subj1,100,70)" represents that subject1 at time step 100 hours had a mean arterial blood pressure between 60-70 mmHg. As noted earlier, not every parameter was measured each hour. The examples were derived from these facts. If on the consideration of two consecutive measurements, there was a significant change in the parameter value (defined as a change of at least two bins in the discretized values), then we generated a positive example. For example, if in addition to the fact listed above, there was "map(subj1,101,80.1)", indicating a significant increase in blood pressure after hour 100, the positive example "mapincr(subj1,100)" would be generated. Otherwise, we synthesized a (false) negative example.

\section{Results}

\begin{figure*}
\centering
  \includegraphics[width=\linewidth]{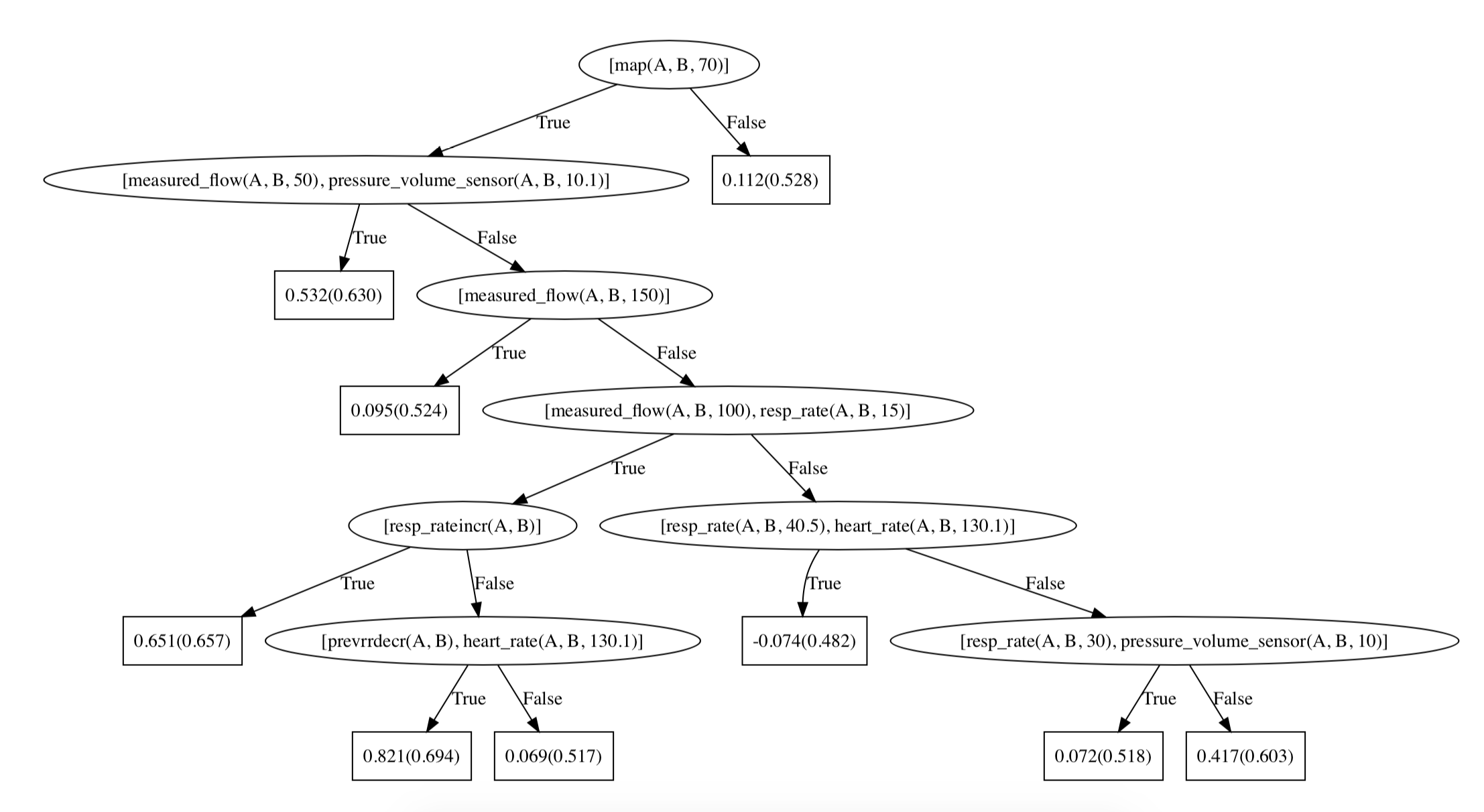}
  \caption{Representative probabilistic logic tree for the action to increase mean arterial pressure (map\_incr). The prevrrdecr(A,B) refers to a decrease in the respiratory rate in the previous hour, expressed by the logic rule \(\mathtt{\exists C \mid B = C+1 \wedge resp\_ratedecr(A,C)}\).}
  \label{fig:tree}
\end{figure*}

In this preliminary experiment, we set the parameters of the boosted learning algorithm so that each concept was approximated by a set of 20 relational regression trees. In these trees, each node consists of a logic clause whose possible truth values are represented by the edges. Leaves of the tree are labelled with the weight (and the value subjected to the \(sigmoid\) function in parentheses, where $sigmoid(x)=1/1+e^{-x}$) corresponding to logic rules constructed by following from the root to the leaf. A representative probabilistic logic tree is presented in Figure~\ref{fig:tree}.  

We extracted weighted first-order logic rules from this tree; the generated clauses are listed in Table~\ref{tab:mapActionClauses}.

%%%%%%%%%%%%%%%%%%%%%%%New table begins
\begin{table*}
%\tiny
\centering
\caption{Weighted first-order logic rules for mean arterial pressure increase generated from a representative boosted tree. Variable A represents the subject. Variables B and C represent the time.}\label{tab:mapActionClauses}
\begin{tabular}{|c|l|l|}
% \begin{comment}
\hline
 No. & Wt. &  \multicolumn{1}{c}{Logic rule}\vline\\
\hline
1 & 0.112 & \( \mathtt{mapincr(A,B)}\,\Leftarrow \mathtt{\neg map(A,B,60-70)}\)\\
2 & 0.532 & \( \mathtt{mapincr(A,B)}\,\Leftarrow \mathtt{map(A,B,60-70)} \wedge \mathtt{measured\_flow(A,B,20-50)} \)\\
& & \( \wedge \mathtt{pressure\_volume\_sensor(A,B,>10)}\)\\
% \begin{comment}
3 & 0.095 & \( \mathtt{mapincr(A,B)}\,\Leftarrow\ \mathtt{map(A,B,60-70)}\wedge \neg [\mathtt{measured\_flow(A,B,20-50)} \) \\
& & \( \wedge \mathtt{pressure\_volume\_sensor(A,B,>10)}]\wedge \mathtt{measured\_flow(A,B,100-150)}\)\\
4 & 0.651 & \(\mathtt{mapincr(A,B)}\,\Leftarrow\ \mathtt{map(A,B,60-70)} \wedge\neg[\mathtt{measured\_flow(A,B,20-50)} \) \\
& & \( \wedge \mathtt{pressure\_volume\_sensor(A,B,>10)}]\wedge\neg\mathtt{measured\_flow(A,B,100-150)} \) \\
& & \( \wedge \mathtt{measured\_flow(A,B,50-100)}\wedge \mathtt{resp\_rate(A,B,\leq15)} \) \\
& & \( \wedge \mathtt{resp\_rateincr(A,B)}\)\\
5 & 0.821 & \(\mathtt{mapincr(A,B)}\,\Leftarrow\ \mathtt{map(A,B,60-70)} \wedge\neg[\mathtt{measured\_flow(A,B,20-50)} \) \\
& &\( \wedge \mathtt{pressure\_volume\_sensor(A,B,>10)}]\wedge\neg\mathtt{measured\_flow(A,B,100-150)} \) \\
& &\( \wedge \mathtt{measured\_flow(A,B,50-100)}\wedge \mathtt{resp\_rate(A,B,\leq15)} \) \\
& & \( \wedge \neg\mathtt{resp\_rateincr(A,B)}\wedge\)
\(\mathtt{heart\_rate(A,B,>130)} \) \\
& & \( \wedge [\mathtt{\exists C \mid B = C+1 \wedge resp\_ratedecr(A,C)}]\)\\
6 & 0.069 & \(\mathtt{mapincr(A,B)}\,\Leftarrow\ \mathtt{map(A,B,60-70)} \wedge\neg[\mathtt{measured\_flow(A,B,20-50)} \) \\
& &\( \wedge \mathtt{pressure\_volume\_sensor(A,B,>10)}]\wedge\neg\mathtt{measured\_flow(A,B,100-150)} \) \\
& &\( \wedge \mathtt{measured\_flow(A,B,50-100)}\wedge \mathtt{resp\_rate(A,B,\leq15)} \) \\
& & \(\wedge \neg\mathtt{resp\_rateincr(A,B)}\wedge\)
\(\neg[\mathtt{heart\_rate(A,B,>130)} \) \\
& &\( \wedge [\mathtt{\exists C \mid B = C+1 \wedge resp\_ratedecr(A,C)}]]\)\\
7 & -0.074 & \(\mathtt{mapincr(A,B)}\,\Leftarrow\ \mathtt{map(A,B,60-70)} \wedge\neg[\mathtt{measured\_flow(A,B,20-50)} \) \\
& &\( \wedge \mathtt{pressure\_volume\_sensor(A,B,>10)}]\wedge\neg\mathtt{measured\_flow(A,B,100-150)} \) \\
& &\( \wedge \neg[\mathtt{measured\_flow(A,B,50-100)}\wedge \mathtt{resp\_rate(A,B,\leq15)}]\)\\
& &\( \wedge \mathtt{resp\_rate(A,B,>40)}\wedge\)
\(\mathtt{heart\_rate(A,B,>130)}\)\\
8 & 0.072 & \(\mathtt{mapincr(A,B)}\,\Leftarrow\ \mathtt{map(A,B,60-70)} \wedge\neg[\mathtt{measured\_flow(A,B,20-50)} \) \\
& & \( \wedge \mathtt{pressure\_volume\_sensor(A,B,>10)}]\wedge\neg\mathtt{measured\_flow(A,B,100-150)} \) \\
& &\( \wedge \neg[\mathtt{measured\_flow(A,B,50-100)}\wedge \mathtt{resp\_rate(A,B,\leq15)}]\)\\
& &\( \wedge \neg[\mathtt{resp\_rate(A,B,>40)}\wedge\)
\(\mathtt{heart\_rate(A,B,>130)}]\)\\
& & \( \wedge \mathtt{resp\_rate(A,B,20-30)}\wedge\)
 \(\mathtt{pressure\_volume\_sensor(A,B,0-10)}\)\\
9 & 0.417 & \(\mathtt{mapincr(A,B)}\,\Leftarrow\ \mathtt{map(A,B,60-70)} \wedge\neg[\mathtt{measured\_flow(A,B,20-50)} \) \\
& & \( \wedge \mathtt{pressure\_volume\_sensor(A,B,>10)}]\wedge\neg\mathtt{measured\_flow(A,B,100-150)} \) \\
& &\( \wedge \neg[\mathtt{measured\_flow(A,B,50-100)}\wedge \mathtt{resp\_rate(A,B,\leq15)}]\)\\
& &\( \wedge \neg[\mathtt{resp\_rate(A,B,>40)}\wedge\)
\(\mathtt{heart\_rate(A,B,>130)}]\)\\
& & \( \wedge \neg[\mathtt{resp\_rate(A,B,20-30)}\wedge\)
 \(\mathtt{pressure\_volume\_sensor(A,B,0-10)}]\)\\
\hline
% \end{comment}
\end{tabular}
\end{table*}

%%%%%%%%%%%%%%%%%%%%%%%%%%New table ends
Owing to the fact that a probabilistic target concept is represented by a sum of the 20 weighted trees, it is difficult to directly interpret the rules generated by our model. However, by comparing some of the weighted rules within a tree, we can elicit findings consistent with known clinical practice. When we look at Rule 1, we see that the weight of the action to increase the mean arterial blood pressure (map) is 0.112 when the map is not between 60-70 mm hg. (which is roughly the normal range). However, comparing to Rule 2, we see that the weight increases to 0.532 when the mean arterial pressure (map) is in the normal range, and if the pump flow is relatively low (20-50 ml/kg-min), and when the pump preload pressure is relatively high ($>10$ cm H2O). The increased weight on this clause, compared to Rule 1, suggests that in circumstances where the map is normal, but if the pump flow is low, physicians may elect to initiate treatment to raise the blood pressure. This is a reasonable treatment maneuver.

We see another example when comparing Rules 5 and 6. The difference between these rules is in the conjunction of the last two clauses:
\begin{multline*}
    \mathtt{heart\_rate(A,B,>130) \, \wedge} \\
    \mathtt{[\exists \, C \mid B = C+1 \wedge resp\_ratedecr(A,C)]}
\end{multline*}
% ($\mathtt{heart\_rate(A,B,>130)})$\wedge
% ([$\mathtt{\exists C \mid B = C+1 \wedge resp\_ratedecr(A,C)}$))].
This clause is present in Rule 5 but negated in Rule 6.
The markedly elevated $heart\_rate$ ($>130$ beats/minute) and a recent decrease in respiratory rate are clinical signs of disease severity. The higher weight on Rule 5 (when these findings are present) indicates that the presence of these findings will result in a higher probability of the physician moving to increase the map, which is clinically very reasonable.

As we add clauses to the rules or negate them, moving down the tree, it is generally true that the changing weights make clinical sense. That is, a physician is able to explain why the rule was created. However, it is also appears that some clauses seem peripheral to the task of deciding whether to increase the arterial pressure, and would not necessarily be used in the clinical setting. Without question, our automatic system is able to generate longer, more complicated probabilistic logic rules and use more (perhaps obscure, and perhaps important) clinical facts than would a human. Whether such rules will be clinically relevant and useful in a functioning clinical decision support system is a question requiring further research. 

\section{Discussion}
We used a statistical relational logic framework to elicit policies for medical management of children undergoing ECMO. Our preliminary results provide some hope that the method can be used to provide interpretable strategies to physicians managing complicated patients that might be useful in automatic clinical decision support systems. To the best of our knowledge, this is the first study using EHR data to learn probabilistic rules governing the management of patients in the hospital. It can be observed from the rules presented that we include existential variables (observations recognized previously); simple encodings into a propositional framework will not suffice for such problems. Instead, a relational framework is necessary. Also, note the weights/regression values on the leaves, demonstrating the need for a ``soft" framework and supports our choice of SRL as a natural choice for such modeling tasks.

There are at least a few shortcomings to our study from a clinical perspective. First, it must be acknowledged that we did not have direct access to the physician actions, and rather derived them from the measured physiologic parameters. This complicates our analysis, owing to the fact that we are unable to distinguish when altered physiologic findings are related to medical care or to the course of the underlying disease. It is reasonable to surmise that when we have available the actual physician orders, we will have a cleaner, less noisy, set of data, perhaps allowing greater success in eliciting the medical policies. A second shortcoming is that in this study we selected only a small subset of the recorded parameters- ones thought to be most physiologically significant in the medical decision-making process. We might obtain better results if we broaden the set of parameters. Moreover, in our SRL experiment, we discovered policies encoded in multiple regression trees; without question, one could question whether the trees learned in the boosting algorithm are readily interpretable to physicians managing patients. This is a broader issue; if weighted logic models such as MLNs/PSL etc., are interepretable, then so are these boosted rules. However, we acknowledge that weighted logic may not be as interpretable to domain experts, and there is a need to explore models that are more explainable and interpretable. 

Finally, our technique may provide new insight into which physician actions contribute to variation in clinical outcome for children undergoing ECMO support. For example, the most common risks of ECMO include bleeding (related to the necessary anticoagulation of the patient) and neurologic injury- either an intracranial hemorrhage or an ischemic event. What is not known is whether neurologic injury risk can be altered by different management schemes. We surmise that when we evaluate policies for patients partitioned by outcome class (that is, with or without neurologic event), aspects of the policies may be elicited that increase complication risk. We leave this to future work. But if such a finding were confirmed in a clinical study, ECMO outcomes could be improved.

\bibliographystyle{aaai}
%
%\begin{thebibliography}{8}
\bibliography{ECMO_AAAI}
%\end{thebibliography}
\end{document}